\documentclass[letterpaper]{article}
\usepackage[margin=1in,dvips]{geometry}
\usepackage{graphicx,psfrag,amsmath,amsthm,amssymb}
\usepackage{natbib}
\usepackage{algorithmic,algorithm}
\usepackage{url}
\usepackage{enumitem}

\newcommand{\dd}{\ensuremath{\mathbf{d}}}

\newcommand{\uu}{\ensuremath{\mathbf{u}}}
\newcommand{\vv}{\ensuremath{\mathbf{v}}}
\newcommand{\w}{\ensuremath{\mathbf{w}}}
\newcommand{\x}{\ensuremath{\mathbf{x}}}
\newcommand{\y}{\ensuremath{\mathbf{y}}}

\newcommand{\0}{\ensuremath{\mathbf{0}}}
\newcommand{\1}{\ensuremath{\mathbf{1}}}

\newcommand{\bbR}{\ensuremath{\mathbb{R}}}

\newcommand{\calL}{\ensuremath{\mathcal{L}}}

\newcommand{\calO}{\ensuremath{\mathcal{O}}}

\newcommand{\abs}[1]{\left\lvert#1\right\rvert}
\newcommand{\norm}[1]{\left\lVert#1\right\rVert}

{%
\begin{list}{#1}{
\vspace{-\topsep}
\vspace{-\partopsep}
\setlength{\itemindent}{0cm}
\setlength{\rightmargin}{0cm}
\setlength{\listparindent}{0cm}
\settowidth{\labelwidth}{#1}
\setlength{\leftmargin}{\labelwidth}
\addtolength{\leftmargin}{\labelsep}
\setlength{\itemsep}{0cm}
}%
}%
{%
\end{list}
\vspace{-\topsep}
\vspace{-\partopsep}
}

{\begin{enumerate}%
}%
{\end{enumerate}}

\theoremstyle{plain}
\newtheorem{thm}{Theorem}[section]

\newtheorem*{lemma*}{Lemma}

\newtheorem*{prop*}{Proposition}

\theoremstyle{definition}

\newtheorem*{defn*}{Definition}

\newtheorem*{exmp*}{Example}

\newtheorem*{conj*}{Conjecture}

\theoremstyle{remark}
\newtheorem{rmk}[thm]{Remark}
\newtheorem*{rmk*}{Remark}

\bibpunct[, ]{(}{)}{;}{a}{,}{,} 

\title{An $\calO(n\log n)$ projection operator for weighted $\ell_1$-norm regularization with sum constraint}
\author{Weiran Wang \\
  Toyota Technological Institute at Chicago \\
  \texttt{weiranwang@ttic.edu}
}
\date{March 1, 2015}

\begin{document}

\maketitle

\begin{abstract}
We provide a simple and efficient algorithm for the projection operator for weighted $\ell_1$-norm regularization subject to a sum constraint, together with an elementary proof. The implementation of the proposed algorithm can be downloaded from the author's homepage.
\end{abstract}

\section{The problem}
In this report, we consider the following optimization problem:
\begin{align} \label{e:proj}
\min_{\x} & \quad \frac{1}{2} \norm{\x-\y}^2 + \sum_{i=1}^n d_i \abs{x_i}, \\
\text{s.t.} & \quad \x^\top \1=1, \nonumber 
\end{align}
where $\y=[y_1,\dots,y_n]^\top \in\bbR^n$, $d_i\ge 0$, $i=1,\dots,n$, and $\1$ is the $n$-dimensional vector consisting of all $1$'s. This is a quadratic program and the objective function is strictly convex (even though it is non-smooth), so there is a unique solution which we denote by $\x=[x_1,\dots,x_n]^\top$ with a slight abuse of notation.

Notice if $d_1=\dots=d_n$ and the constraint were absent, the problem has a closed form solution known as the soft-shrinkage operator (see, e.g., \citealp{BeckTeboul09a}), which is widely used for solving $\ell_1$-regularized problem in learning sparse representations. But our problem \eqref{e:proj} is more involved due to the constraint that couples all dimensions of $\x$. Nonetheless, we give an efficient algorithm with time complexity $\calO(n\log n)$ for this problem using only the KKT theorem.

\begin{rmk}
Our motivation for \eqref{e:proj} also comes from sparse coding. \citet{Yu_09a} propose the local coordinate coding (LCC) algorithm for learning sparse representations induced by locality. Given a data sample $\uu\in \bbR^n$ and a set of landmark points $\{\vv_j\}_{j=1}^C$ where $\vv_j \in \bbR^n$, $j=1,\dots,C$, the LCC algorithm reconstructs $\uu$ from the landmark points while enforcing the faraway landmark points to contribute less than nearby landmark points (or to have smaller reconstruction coefficients). Let the reconstruction coefficient of $\vv_j$ be $w_j$, $j=1,\dots,C$. Then the optimization problem for these coefficients in LCC is
\begin{align}\label{e:lcc}
\min_{\w} & \quad \norm{\uu-\sum_{j=1}^C w_j \vv_j}^2 + \lambda \sum_{j=1}^C {\norm{\uu-\vv_j}^2 \abs{w_j} } \\
\text{s.t.} & \quad \sum_{j=1}^C w_j=1, \nonumber
\end{align}
where  $\lambda>0$ is some trade-off parameter. The constraint in \eqref{e:lcc}  ensures that the representation is translation invariant. There are different ways of solving this problem, e.g., \citet{ElhamifVidal11a} have a similar optimization problem which they solve with Alternating Direction Method of Multipliers \citep{Boyd_11a}. One simple way of solving \eqref{e:lcc} is to use the gradient proximal algorithm and its Nesterov's acceleration scheme (see \citealp{BeckTeboul09a} and the reference therein), where one iteratively takes a short gradient step for the smooth quadratic term and projects the new estimate with the weighted $\ell_1$ regularization term subject to the sum constraint, where the projection operator solves exactly \eqref{e:proj}.
\end{rmk}

\section{The solution}
We solve the problem \eqref{e:proj} using only the KKT theorem \citep{NocedalWright06a}, which states the necessary and sufficient condition\footnote{Strictly speaking, our objective is convex and  non-smooth, so the condition is that the zero vector $\0$ lies in the sub-differential at the solution $\x$.} satisfied by the solution $\x$. The Lagrangian of \eqref{e:proj} is
\begin{equation}
\calL(\x,\alpha)= \frac{1}{2} \norm{\x-\y}^2 +  \sum_{i=1}^n d_i \abs{x_i}+\alpha (\x^\top \1-1),
\end{equation}
where $\alpha$ is the Lagrange multipliers associated with the constraint. And the KKT system of this problem is
\begin{subequations} \label{e:kkt}
\begin{align}
x_i-y_i+ d_i+\alpha=0, &\qquad \text{if}\quad x_i>0,\\
x_i-y_i- d_i+\alpha=0, &\qquad \text{if}\quad x_i<0,\\
- d_i \le -y_i+\alpha \le  d_i, &\qquad \text{if}\quad x_i=0, \label{e:abs}\\
\sum_{i=1}^n x_i=1,
\end{align}
\end{subequations}
where we have used the fact that the sub-differential of $\abs{x}$ is $[-1,1]$ at $x=0$ to obtain \eqref{e:abs}.

Denote $y_i^- = y_i- d_i$, $y_i^+ = y_i+ d_i$, $i=1,\dots,n$, which can be computed beforehand. We can then rewrite \eqref{e:kkt} in terms of $\alpha$:
\begin{subequations}\label{e:alpha}
\begin{align}
\alpha < y_i^- \; \Longleftrightarrow x_i>0, \label{e:alpha1}\\
\alpha > y_i^+ \; \Longleftrightarrow x_i<0, \label{e:alpha2}\\
y_i^- \le \alpha\le y_i^+ \; \Longleftrightarrow x_i=0, \label{e:alpha3}\\
\sum_{i:\; x_i>0} (y_i^- -\alpha) + \sum_{i:\; x_i<0} (y_i^+ -\alpha) =1. \label{e:alpha4}
\end{align}
\end{subequations}

Obviously, the Lagrange multiplier $\alpha$ is the key to our problem. Once the value of $\alpha$ is determined, we can easily obtain the optimal solution by setting
\begin{subequations}\label{e:sol}
\begin{align}
x_i &= y_i^- -\alpha \qquad   \text{if}\;\; y_i^-  > \alpha, \\
x_i &= y_i^+ -\alpha \qquad  \text{if}\;\; y_i^+  < \alpha, \\
x_i &= 0  \qquad\qquad\ \   \text{otherwise}.
\end{align}
\end{subequations}
We can sort all dimensions of $y_i^-$ and $y_i^+$ together (a total of $2N$ scalars) into an ascending $z$-sequence:
\begin{equation}
z_1 \le z_2 \le \cdots \le z_{2N}.
\end{equation}
An important observation is that the $z$-sequence partitions the real axis into $4N+1$ disjoint sets, each being either a single point set $\{z_j\}$, $j=1,\dots,2N$ or an open interval of the form $(-\infty,z_1)$, $(z_j,z_{j+1})$, $j=1,\dots,2N-1$, or $(z_{2N},\infty)$ and the Lagrange multiplier $\alpha$ for the solution must lie in one of them. 

We then test each of the $4N+1$ sets as follows. Assuming that $\alpha$ lies in one set, we can use \eqref{e:alpha1}--\eqref{e:alpha3} to conjecture the positive, negative, and zero dimensions of a possible solution $\hat{\x}$. After that, we use \eqref{e:alpha4} to compute a hypothesized value $\hat{\alpha}$ for the Lagrange multiplier, i.e., 
\begin{align}
\alpha=\frac{\sum\limits_{i:\; \hat{x}_i>0} y_i^- + \sum\limits_{i:\; \hat{x}_i<0} y_i^+ -1}{\sum\limits_{i:\; \hat{x}_i>0} 1 + \sum\limits_{i:\; \hat{x}_i<0} 1 }.
\end{align}
If the computed $\hat{\alpha}$ indeed lies in the assumed set (a point or an open interval), we have a KKT point and thus the solution. 

Since the problem \eqref{e:proj} is strictly convex and there exists a unique global optimum, this procedure will find the exact solution with no more than $4N+1$ tests. We can do this efficiently by sorting $y_i^-$ and $y_i^+$ separately ($\calO(n\log n)$ operations) and gradually merging the two sorted sequences (an $\calO(n)$ operation). Therefore the total cost of our procedure for solving \eqref{e:proj} is of order $\calO(n\log n)$.

Algorithm~\ref{alg:proj} gives the detailed pseudocode for solving \eqref{e:proj}, whose MATLAB and \texttt{C++} implementation can be downloaded at \texttt{https://eng.ucmerced.edu/people/wwang5}.

\begin{algorithm}
  \caption{Pseudo-code of our projection operator for \eqref{e:proj}.}
  \label{alg:proj}\small
  \renewcommand{\algorithmicrequire}{\textbf{Input:}}
  \renewcommand{\algorithmicensure}{\textbf{Output:}}
  \begin{algorithmic}
    \REQUIRE $\y \in\bbR^n$ and $\dd=[d_1,\dots,d_n]$ where $d_i\ge 0$, $i=1,\dots,n$.
    \STATE Sort $\y- \dd$ into $\y^-$: $y_1^- \le y_2^- \le \dots \le y_n^-$.\ \ And sort $\y+ \dd$ into $\y^+$: $y_1^+ \le y_2^+ \le \dots \le y_n^+$.
    \STATE $i\leftarrow 1$,\ \ $j\leftarrow 1$ \hfill\textsf{\% $i/j$ index of the dimension of $\y^-/\y^+$ that will be merged next.}
    \STATE \textsf{\% $s_1/s_2$ stores the sum of dimensions of $\y^-/\y^+$  that are strictly greater/smaller than the current estimate of $\alpha$.}
    \STATE $s_1 \leftarrow \sum_{i=1}^n y_i^-$,\ \ $s_2 \leftarrow 0$,\ \ $t\leftarrow n$ \hfill\textsf{\% $t$ is the number of nonzero dimensions of the hypothesized $\x$.}
    \IF {$(s1+s2)<t\cdot y_1^-$} 
    \STATE $\alpha\leftarrow (s1+s2)/t$; \textbf{return}  \hfill\textsf{\% $\alpha<y_1^-$, all dimensions of $\x$ are positive.} 
    
    \ENDIF
    \WHILE{\textbf{true}} 
    \STATE \textsf{\% Test a single point set.}
    \IF {$y_i^- < y_j^+$}               
    \STATE $k\leftarrow i$  \hfill\textsf{\% $y_i^-$ is the next value in the $z$-sequence.}
    \WHILE {$(y_k^-=y_i^-) \ \&\&\  (k\le n)$}
    \STATE $s_1\leftarrow s_1-y_k^-$,  $t\leftarrow t-1$, $k\leftarrow k+1$ \hfill\textsf{\% Skip the contiguous block of identical dimensions in $\y^-$.}
    \ENDWHILE
    \IF {$(s_1+s_2-1)=t \cdot y_i^-$} 
    \STATE $\alpha\leftarrow y_i^-$; \textbf{return} \hfill\textsf{\% $\alpha$ happens to lie in a single point set.}
    \ELSE 
    \STATE $left\leftarrow y_i^-$, $i\leftarrow k$ \hfill\textsf{\% Otherwise, $\alpha$ lies in a open interval with left boundary $left$.}
    \ENDIF
    \ELSE \IF {$y_i^- > y_j^+$}
    \STATE \textsf{\% $y_j^+$ is the next value in the $z$-sequence.}
    \IF {$(s_1+s_2-1)=t \cdot y_j^+$} 
    \STATE $\alpha\leftarrow y_j^+$; \textbf{return} \hfill\textsf{\% $\alpha$ happens to lie in a single point set.}
    \ELSE 
    \STATE $left\leftarrow  y_j^+$ \hfill\textsf{\% Otherwise, $\alpha$ lies in a open interval with left boundary $left$.}
    \WHILE {$(y_j^+=left) \ \&\&\  (j\le n)$}
    \STATE $s_2\leftarrow s_2+y_j^+$,  $t\leftarrow t+1$, $j\leftarrow j+1$ \hfill\textsf{\% Skip the contiguous block of identical entries in $\y^+$.}
    \ENDWHILE
    \ENDIF
    \ELSE
    \STATE $k\leftarrow i$ \hfill\textsf{\% $y_i^-=y_j^+$ is the next value in the $z$-sequence.}
    \WHILE {$(y_k^-=y_i^-) \ \&\&\  (k\le n)$}
    \STATE $s_1\leftarrow s_1-y_k^-$,  $t\leftarrow t-1$, $k\leftarrow k+1$ 
    \ENDWHILE
    \IF {$(s_1+s_2-1)=t \cdot y_i^-$} 
    \STATE $\alpha\leftarrow y_i^-$; \textbf{return}
    \ELSE 
    \STATE $left\leftarrow  y_i^-$, $i\leftarrow k$
    \WHILE {$(y_j^+=left) \ \&\&\  (j\le n)$}
    \STATE $s_2\leftarrow s_2+y_j^+$,  $t\leftarrow t+1$, $j\leftarrow j+1$ 
    \ENDWHILE
    \ENDIF
    \ENDIF
    \ENDIF
    \STATE \textsf{\% Find the right boundary of the open interval and test if it contains $\alpha.$}
    \IF {$y_i^- < y_j^+$}
    \STATE $right\leftarrow y_i^-$
    \ELSE
    \STATE $right\leftarrow y_j^+$
    \ENDIF
    \IF {$t\cdot left<(s_1+s_2-1)\ \&\&\  t\cdot right > (s_1+s_2-1)$} 
    \STATE $\alpha\leftarrow (s_1+s_2-1)/t$;  \textbf{return} \hfill\textsf{\% $\alpha$ lies in the open interval $(left, right)$.}
    \ENDIF
    \ENDWHILE
    \ENSURE $\alpha$ is the Lagrange multiplier of the problem \eqref{e:proj}, use \eqref{e:sol} to obtain $\x$.
  \end{algorithmic}
\end{algorithm}

\bibliographystyle{abbrvnat}

\begin{thebibliography}{5}
\providecommand{\natexlab}[1]{#1}
\providecommand{\url}[1]{\texttt{#1}}
\expandafter\ifx\csname urlstyle\endcsname\relax
  \providecommand{\doi}[1]{doi: #1}\else
  \providecommand{\doi}{doi: \begingroup \urlstyle{rm}\Url}\fi

\bibitem[Beck and Teboulle(2009)]{BeckTeboul09a}
A.~Beck and M.~Teboulle.
\newblock A fast iterative shrinkage-thresholding algorithm for linear inverse
  problems.
\newblock \emph{SIAM J. Imaging Sciences}, 2\penalty0 (1):\penalty0 183--202,
  2009.

\bibitem[Boyd et~al.(2011)Boyd, Parikh, Chu, Peleato, and Eckstein]{Boyd_11a}
S.~Boyd, N.~Parikh, E.~Chu, B.~Peleato, and J.~Eckstein.
\newblock Distributed optimization and statistical learning via the alternating
  direction method of multipliers.
\newblock \emph{Foundations and Trends in Machine Learning}, 3\penalty0
  (1):\penalty0 1--122, 2011.

\bibitem[Elhamifar and Vidal(2011)]{ElhamifVidal11a}
E.~Elhamifar and R.~Vidal.
\newblock Sparse manifold clustering and embedding.
\newblock In J.~Shawe-Taylor, R.~S. Zemel, P.~Bartlett, F.~Pereira, and K.~Q.
  Weinberger, editors, \emph{Advances in Neural Information Processing Systems
  (NIPS)}, volume~24, pages 55--63. MIT Press, Cambridge, MA, 2011.

\bibitem[Nocedal and Wright(2006)]{NocedalWright06a}
J.~Nocedal and S.~J. Wright.
\newblock \emph{Numerical Optimization}.
\newblock Springer Series in Operations Research and Financial Engineering.
  Springer-Verlag, New York, second edition, 2006.

\bibitem[Yu et~al.(2009)Yu, Zhang, and Gong]{Yu_09a}
K.~Yu, T.~Zhang, and Y.~Gong.
\newblock Nonlinear learning using local coordinate coding.
\newblock In Y.~Bengio, D.~Schuurmans, J.~Lafferty, C.~K.~I. Williams, and
  A.~Culotta, editors, \emph{Advances in Neural Information Processing Systems
  (NIPS)}, volume~22. MIT Press, Cambridge, MA, 2009.

\end{thebibliography}

\end{document}